%% file: main.tex
\documentclass[runningheads]{llncs}

\usepackage[mobile]{eccv} % It is fine.

\usepackage{graphicx}
\usepackage{booktabs}

\newif\ifarxiv
\arxivtrue
\ifarxiv
\usepackage[pagebackref=false,breaklinks=true,colorlinks,citecolor=eccvblue,bookmarks=true,bookmarksnumbered=true]{hyperref}
\hypersetup{
  pdftitle={ES-PTAM: Event-based Stereo Parallel Tracking and Mapping},
  pdfsubject={Computer Vision, Robotics},
  pdfauthor={Suman Ghosh, Valentina Cavinato, Guillermo Gallego},
  pdfkeywords={SLAM, ego-motion estimation, Event Cameras, Asynchronous Sensor, High Dynamic Range, High Temporal Resolution}
}
\usepackage[absolute]{textpos}
\else
\usepackage{hyperref}
\fi

\usepackage{orcidlink} % Support for ORCID icon
\usepackage{amsfonts}
\usepackage{enumitem}
\usepackage{siunitx}
\DeclareSIUnit{\nothing}{\relax}
\usepackage{xcolor}
\usepackage{pifont}

\usepackage{makecell}
\usepackage{adjustbox} % resize width if needed
\usepackage{multirow}
\usepackage{mathrsfs}
\usepackage{empheq}
\usepackage{rotating}

\input{macros}
\begin{document}
\title{ES-PTAM: Event-based Stereo Parallel\\ Tracking and Mapping}

% arXiv
\ifarxiv
\definecolor{somegray}{gray}{0.5}
\newcommand{\darkgrayed}[1]{\textcolor{somegray}{#1}}
\begin{textblock}{13}(1.5, -0.1)  % {hsize}(hpos,vpos)
\begin{center}
\darkgrayed{This paper has been accepted for publication at the
European Conference on Computer Vision (ECCV) Workshops, Milan, Italy, 2024.
\copyright Springer
}
\end{center}
\end{textblock}
\fi

\author{Suman Ghosh\inst{1,2}\orcidlink{0000-0002-4297-7544} \and
Valentina Cavinato\inst{3}\orcidlink{0009-0007-4500-5592} \and
Guillermo Gallego\inst{1,2,4,5}\orcidlink{0000-0002-2672-9241}
}

\authorrunning{S.~Ghosh et al.}

\institute{Technical University of Berlin, Berlin, Germany \and 
Robotics Institute Germany, Berlin, Germany \and
Sony Semiconductor Solutions Europe, Sony Europe B.V., \\ Stuttgart Laboratory 1, Zürich Office, Switzerland \and
Science of Intelligence Excellence Cluster, Berlin, Germany \and 
Einstein Center Digital Future, Berlin, Germany
}

\maketitle

%%%%%%%%% ABSTRACT
\begin{abstract}
% % How to write an abstract? 
% % Slides 16 onward from: Shortcomings in scientific writing by Jean Luc Doumont
% % Context: What you need to know to understand the need
Visual Odometry (VO) and SLAM are fundamental components for spatial perception in mobile robots.
%in order for them to understand their environment and interact with it.
% % Need (= problem): A way to motivate the audience
Despite enormous progress in the field, current VO/SLAM systems are limited by their sensors' capability.
Event cameras are novel visual sensors that offer advantages to overcome the limitations of standard cameras, 
enabling robots to expand their operating range to challenging scenarios, such as high-speed motion and high dynamic range illumination. 
% % Task: What I decided or was asked to do (in order to address the need)
% % Object: What the document does/contains (and what you should do with it)
We propose a novel event-based stereo VO system by combining two ideas: 
a correspondence-free mapping module that estimates depth by maximizing ray density fusion 
and a tracking module that estimates camera poses by maximizing edge-map alignment.
% % Findings: What I found as a result of carrying out the task
We evaluate the system comprehensively on five real-world datasets, spanning a variety of camera types (manufacturers and spatial resolutions) and scenarios (driving, flying drone, hand-held, egocentric, etc).
The quantitative and qualitative results demonstrate that our method outperforms the state of the art in majority of the test sequences by a margin,
e.g., trajectory error reduction of 45\% on RPG dataset, 61\% on DSEC dataset, and 21\% on TUM-VIE dataset.
% % Conclusion (= interpretation/recommendation): What the findings mean to you (possibly what you should do). Give concrete numbers.
% % Perspectives: What I/you/they will do in the future
% We also discuss the limitations of the method and hope that our work inspires further research to endow machines with more capable visual perception systems based on event cameras.
To benefit the community and foster research on event-based perception systems, we release the source code and results: \url{https://github.com/tub-rip/ES-PTAM}.
\end{abstract}

\section{Introduction}
\label{sec:intro}

Visual odometry (VO) and simultaneous localization and mapping (SLAM) are essential techniques for solving the task of perceiving a robot's location and its 3D environment from visual inputs. 
This task presents significant challenges, especially in high-speed motion and high dynamic range (HDR) illumination conditions, 
where conventional cameras may fail due to a common exposure time for all pixels, which limits dynamic range and may produce motion blur. 
The need for onboard processing in robots with limited computational resources further adds to these challenges \cite{Cadena16tro}.
Event cameras are novel bio-inspired sensors that address these challenges by detecting per-pixel brightness changes asynchronously at a high temporal resolution ($\mu$s), offering low power consumption, high dynamic range and sparsity for robust and efficient VO/SLAM \cite{Lichtsteiner08ssc,Gallego20pami,Gehrig24nature}.

Progress in event-based VO/SLAM has been witnessed in monocular \cite{Kim16eccv,Rebecq17ral,Klenk24threedv},
and stereo setups \cite{Hadviger21ecmr,Wang23ral,Zhu18eccv,Zhou20tro,Elmoudni23itsc,Ghosh22aisy},
possibly aided by other sensors (e.g., depth and/or IMU sensor fusion) \cite{Rosinol18ral,Hidalgo22cvpr,Zou22icra}.
We focus on the stereo case (\cref{fig:eyecatcher}), which is particularly suitable for automotive applications, supported by the introduction of new datasets \cite{Zhu18ral,Gehrig21ral,Chaney23cvprw,Mollica13tits}, 
because it can recover the absolute scale of the scene 
and produces fast depth estimates due to spatial parallax.
Moreover, we consider the event-only setting, as our objective is to advance knowledge in purely event-based processing systems, expanding their capabilities while avoiding potential bottlenecks associated with paired sensors \cite{Klenk24threedv}. 

Event-based stereo VO/SLAM methods can be categorized according to multiple criteria (see \cref{tab:relworks}).
Similar to the classification employed for frame-based systems \cite{Engel17pami}, they can be either \emph{feature-based} (indirect) \cite{Hadviger21ecmr,Wang23ral} or \emph{direct} \cite{Zhou20tro,Elmoudni23itsc,Ghosh22aisy}.
Feature-based methods convert the event data produced by the camera into features (e.g., keypoints), thereby converting the VO/SLAM problem into one based on multi-view geometry. 
This enables the utilization of  classical solutions.
Besides compatibility, summarizing the event stream into a few local features also offers speed-up benefits. 
However, detecting accurate and stable feature tracks poses a challenge due to the dependency of events on motion (unlike conventional images that have motion-invariant properties) and the high amounts of noise (due to the pixels operating at low power). 
These factors undermine the effectiveness of feature-based VO/SLAM methods.
By contrast, direct methods utilize all events recorded by the camera rather than relying solely on those that meet the criteria for a feature. 
As events are triggered by moving edges, these methods often lead to semi-dense approaches (e.g., involving alignment and/or recovery of edge-like structures). 
\input{floats/fig_mapping_tumvie}

\input{floats/tab_related_works}

In this paper, we adopt a direct approach, motivated by the fact that direct methods are state-of-the-art in many event-based motion estimation tasks \cite{Guo24tro,Nunes21pami}.
Specifically, we improve upon the state-of-the-art mapping method (Multi-Camera Event-based Multi-View Stereo -- MC-EMVS) \cite{Ghosh22aisy}, 
capable of generating semi-dense depth maps from synchronized event cameras. 
We combine this method with a simple, fast and effective event camera tracker grounded on the idea of (semi-dense) edge-map alignment \cite{Rebecq17ral}.

In contrast to the pioneering work Event-based Stereo Visual Odometry (ESVO) \cite{Zhou20tro}, the MC-EMVS mapper offers several advantages, such as sharper edges, higher accuracy and improved depth map completion \cite{Ghosh22aisy}. 
We aim to capitalize on these benefits within a parallel tracking-and-mapping system.
Our contributions pertain to system-level aspects, 
focusing first on designing a coherent system (with compatible mapper and tracker) 
and then on optimizing performance, thus blending the goals of each module into the central objective of accurate VO.
We evaluate our approach on multiple datasets, with different event-camera types, spatial resolutions, calibrations and scene geometries. 
\Cref{tab:relworks} summarizes the theoretical comparison among the state of the art in this topic, as well as the datasets on which the methods have been tested.

As \cref{tab:relworks} shows, while MC-EMVS has been used in El Moudni et al.~\cite{Elmoudni23itsc}, the latter uses different underlying principles for the representations in the tracker and the mapper (time surfaces (TS) vs event counts, respectively), which makes it discordant. 
The reported results are also limited (only mapping results on a single dataset).
Moreover, time surfaces, borrowed from ESVO, define an anisotropic distance field, which results in a biased preference for certain motions. 
By contrast, our tracker considers an isotropic distance on the image plane, which does not suffer from such a bias. We also characterize overall tracking and mapping performance of the full system with a comprehensive evaluation.

Our work deviates from Zhu et al.~\cite{Zhu19cvpr}, which proposes a supervised learning-based system that is trained on stereo data to predict ego-motion and depth, but that at inference time works only on monocular setups. 
We also differ from Shiba et al.~\cite{Shiba24pami}, which parameterized the problem using a depth map and optimizes for optical flow, ego-motion and scene depth. 
Their method only works for short time intervals (where the optical flow is a sensible motion model) and provides proof-of-concept results on stereo data. 
By contrast, our method works on larger (arbitrary) observation intervals, since we do not use the linear motion approximation (optical flow), 
but rather the full non-linear multi-view equations. 

Our main \textbf{contributions} can be summarized as follows:
\begin{enumerate}[noitemsep,nolistsep]
    \item A novel event-only stereo VO/SLAM system, which combines an improved version of the MC-EMVS mapper and an edge alignment-based tracker. 
    To the best of our knowledge, it is the first time that this system is proposed. 
    \item A comprehensive evaluation on \emph{five} datasets, more than any other stereo VO method, 
    outperforming the state of the art in most test sequences.
    \item An event-based VO system that naturally scales to multi-camera ($\geq 2$) setups thanks to the fusion-based architecture of the mapping module. 
    We show, for the first time, results on event-based trinocular VO/SLAM.
     \item The source code and results of our stereo system.
\end{enumerate}

\section{Methodology}
\label{sec:method}
\input{floats/fig_pipeline}

\Cref{fig:pipeline} provides an overview of the proposed system. 
It consists of two main blocks (tracking and mapping modules) operating in parallel (at different rates).
Both modules receive events as inputs and rely on each other's output to maintain proper functionality.

The mapping module is based on MC-EMVS \cite{Ghosh22aisy},
which demonstrated the state-of-the-art depth estimation capabilities when using ground truth (GT) poses. 
This work proposes additional improvements and demonstrates that the resulting system yields state-of-the-art performance even when pose accuracy is compromised, as shown by its interaction with a simple event camera tracker.

The tracking module is derived from EVO \cite{Rebecq17ral}, which is grounded on the idea of global image alignment \cite{Baker04ijcv}.
Whereas the map in EVO \cite{Rebecq17ral} is given by EMVS \cite{Rebecq18ijcv}, our depth estimates come from a mapping module which
extends \cite{Rebecq18ijcv} to multi-camera data fusion via the harmonic mean. 
The upcoming sections describe each module after briefly reviewing the event camera's working principle.

\subsection{Event Camera Working Principle}
\label{sec:method:eventcam}
Event cameras, such as the Dynamic Vision Sensor (DVS) \cite{Lichtsteiner08ssc}, 
transmit pixel-wise brightness changes asynchronously, in the form of ``events''.
An event $e_k \doteq (\bx_k, t_k, \pol_{k})$ is produced when the change in log-brightness $\Lum$ reaches a threshold $\theta>0$:
\begin{equation}
\label{eq:generativeEventCondition}
\Lum(\bx_k,t_k) - \Lum(\bx_k, t_k-\Delta t_k) = \pol_k \, \theta,
\end{equation} 
where $\bx_k\doteq (x_k, y_k)^{\top}$, $t_k$ (in \si{\micro\second}) and $\pol_{k} \in \{+1,-1\}$
are the space-time coordinates and polarity of the event, respectively,
and $t_k-\Delta t_k$ is the time of the previous event at the same pixel $\bx_k$.
Assuming constant illumination, events are triggered by moving edges on the image plane.

\subsection{Stereo Mapping by Ray Density Fusion} 
\label{sec:method:mapper}
Our mapper works on the principle of ray density fusion across stereo cameras \cite{Ghosh22aisy}. 
Events are back-projected through 3D space at their corresponding camera location by shooting fictitious rays through 3D space following a space-sweep approach \cite{Collins96cvpr}.
These give rise to intermediate volumetric representations called Disparity Space Images (DSI),
where each point in the volume represents the count of rays that pass through it. 
Hence, the DSI from each camera reflects the density of viewing rays originating from the events; 
the higher the density, the higher the likelihood of that point being the location of a 3D edge that triggers the events.
The DSIs built from the stereo cameras are then fused into a merged DSI that incorporates the parallax from the stereo configuration and balances the events triggered at each camera. 
This fused DSI is then used to extract depth information (candidate locations of 3D edges) by local maxima detection.

Formally, let $\cE_l=\{e^l_{k}\}_{k=1}^{\numEvents^l}$ ($l\equiv$ left) and $\cE_r=\{e^r_{k}\}_{k=1}^{\numEvents^l}$ ($r\equiv$ right) be stereo events over some time interval $[0,T]$, 
and ${f}_l,{f}_r:V\subset\mathbb{R}^3 \to \mathbb{R}_{\geq 0}$ be the corresponding ray densities (DSIs) defined over a volume $V$. 
That is, 
\begin{equation}
\label{eq:LeftDSI}
\textstyle
{f}_l(\bX) = \sum_{k=1}^{\numEvents^l}\delta\bigl(\bX-\bX'_k(e^l_{k})\bigr),
\end{equation}
where $\bX'_k(e^l_{k}) = (\bx^{l\prime \top}_k,Z)^\top$ is a 3D point on the back-projected ray through event $e^l_{k}$, at depth $Z$ with respect to a reference view (RV). 
Events are transferred to RV using a warp $\Warp$ defined by the continuous motion of the cameras and candidate space-sweep depth values $Z\in [Z_{\min},Z_{\max}]$,  
$\label{eq:WarpThatTransfersPointsTwoCameras}\bx^{l\prime}_k = \Warp\bigl(e^l_{k}, \mP^l(t_k),\mP_{v}, Z \bigr),$
where $\mP^l(t)$ is the pose of the left event camera at time $t$ and $\mP_v$ is the pose of RV.
In implementation, DSIs are discretized over a projective voxel grid with $N_Z$ depth planes, 
and the Delta $\delta$ in \eqref{eq:LeftDSI} is approximated by bilinear voting \cite{Rebecq18ijcv}.
Hence, each voxel counts the number of event-viewing rays that pass through~it.

Among the multiple fusion schemes available in MC-EMVS we choose a point-wise (i.e., voxel-wise) harmonic mean fusion across cameras and no extra fusion in time due to its high accuracy and speed \cite{Ghosh22aisy}.
Specifically, if $f_l(\bX), f_r(\bX)$ are the DSI values at a 3D point $\bX$, the fused DSI value at that point is given by
\begin{equation}
\label{eq:DSIfusion}
f_\text{stereo}(\bX) = 2\, /\, (1/f_l(\bX)+1/f_r(\bX)).
\end{equation}
The output depth $Z^\ast$ at each pixel of the RV is obtained from the fused DSI as the local maxima after applying adaptive Gaussian thresholding and a median filter for denoising. 
This approach has connections with Contrast Maximization \cite{Gallego18cvpr}.
The maximum of the DSI at each of its viewing pixels allows us to compute the ``confidence map'' (\cref{fig:eyecatcher}); 
the higher its value the more rays intersect from both cameras, providing more evidence for the presence of scene edges
(note that \cref{fig:eyecatcher} is in negated form: dark indicating high confidence).
Finally, the depth map is converted into a point cloud (local 3D map $\mathcal{M}$ at RV).

\subsubsection{Improvements with respect to MC-EMVS \cite{Ghosh22aisy}.}
The mapper processes events in batches of fixed number of events $N_e$ (instead of a constant time window \cite{Ghosh22aisy}). 
This adapts better to the amount of motion in the scene assuming that the texture is approximately constant.
Also, for improved robustness to noise, we set a minimum and maximum limit for the duration spanned by $N_e$ events, 
to deal with cases with too high event rate (e.g., flashing lights) or too low event rate (e.g., ambient noise while stationary).

\input{floats/fig_ray_improvements}
Additionally, we propose a new event back-projection strategy to deal with scenarios where the camera trajectory lies within the 3D volume spanned by the current DSI. This happens, for instance, when the camera moves along its optical axis, in the forward or backward direction. 
In such scenarios, a pathologically high ray density is produced in the DSI due to the fact that the rays emanate from points inside the DSI (i.e., the optical centers of the camera poses).
The optical centers of other camera locations seen from the reference view (RV) thus erroneously appear as highly confident 3D point estimates.
This happens because all the rays that cast from a camera also seem to intersect at its own optical center.
MC-EMVS \cite{Ghosh22aisy} circumvented this issue by always setting the DSI RV at the end of the camera trajectory in forward moving scenes, but this strategy does not work for arbitrary motions (e.g., moving backwards).
Here, we provide a general solution to the problem for all kinds of motion by limiting the ray casting to depth planes beyond $Z_{\min}$ distance from the optical center in the forward direction. 

\Cref{fig:rayimprovements} compares the proposed strategy to that in MC-EMVS \cite{Ghosh22aisy} on a section of the DSEC driving dataset \cite{Gehrig21ral}, showing a clear improvement. 
This modification allows us to unlock 3D mapping even during forward and backward motion along the camera's optical axis, which is necessary for robust \mbox{6-DOF} VO.

\subsection{Camera Tracking by Edge-Map Alignment}

Our system tracks the motion of the stereo rig using only one of the cameras (``Left'' in \cref{fig:pipeline}).
This approach is chosen because tracking using more than one camera does not offer significant advantages, but it introduces additional computational cost (see \cite{Zhou20tro}).

Since we aim to demonstrate the effectiveness of the fusion-based mapping module for visual odometry, 
we use an event-based camera tracker that works on the produced semi-dense 3D maps, by leveraging the idea of edge-map alignment \cite{Rebecq17ral}.
This tracker operates by minimizing (in the Lie group of rigid body motions $SE(3)$) a simplified photometric error between two bimodal images. 
Specifically, 
\begin{equation}
    \label{eq:treacker:loss}
    \textstyle
    \min_{T\in SE(3)} \sum_{\bp} \bigl(E(\bp; T, \mathcal{M}) - B(\bp) \bigr)^2.
\end{equation}

One image is an edge map $E$ obtained by projecting the current semi-dense 3D map (point cloud) $\mathcal{M}$ onto the camera’s image plane at a candidate pose $T\in SE(3)$. 
The other image is formed by event counting: as the camera moves, a small number of events are accumulated into a binary image $B$ (pixels where events have been triggered are marked as ‘1’ over a ‘0’ canvas). 
If the number of events is small, there is no motion blur in the accumulated image, i.e., it forms a sharp edge map that resembles the projected point cloud. 
Consequently, we estimate the incremental pose necessary to align the projected map with the event image by minimizing their ``intensity'' error \eqref{eq:treacker:loss} (sampled at a subset of the pixels, for speed-up). 
The minimization is achieved using the inverse compositional Lucas-Kanade algorithm \cite{Baker04ijcv}. 
To aid the optimizer, smoothness in the error function is introduced by applying a Gaussian blur to the projected map.

\input{floats/tab_metrics_all}

\subsection{Module Interaction: Parallel Tracking and Mapping}
The mapping and tracking modules described in the previous sections operate in parallel, communicating (\cref{fig:pipeline}) via ROS \cite{Quigley09icraoss}.
Both modules operate asynchronously in an event-driven manner.
During motion, camera poses are progressively estimated (every $M$ events) by the tracker (by composing incremental poses) as long as there is sufficient overlap in the field of view (FOV) between the already estimated 3D maps and the current camera view. 
When the camera moves into an unobserved part of the scene that has not yet been mapped or when the motion parallax is above a threshold, a new map update is requested using all the estimated poses and the most recent $N_e$ events from both cameras.

In addition to the above ``on-demand'' mode of operation (FOV overlap), it is also possible to operate the mapper at a fixed rate.
Since the mapping process is computationally more expensive than the tracking operation, 
it is done at a considerably slower rate (e.g., 5 maps/s vs. 50--150 poses/s). 
It is advisable to run the mapper in ``on-demand'' mode for scenarios with large event rates like with high-speed motion or high camera resolution.
In practice, for efficiency and memory optimization, we retain only a single local map for pose estimation, which has proven to be sufficient for proper operation. 
Local maps are aggregated into a global point cloud for visualization.

There are several ways in which the system can be initialized / bootstrapped, depending on the specific event camera used.
One possibility is, as in \cite{Rebecq17ral}, initialization by classical epipolar-geometry methods on event images.
Some devices, like the DAVIS \cite{Taverni18tcsii}, concurrently output events, grayscale frames and IMU data.
In such cases, initial poses may be obtained from running a frame-based method (e.g., SVO \cite{Forster17troSVO}) on the DAVIS frames, 
by IMU dead-reckoning or by a VIO method (e.g., EVIO \cite{Zhu17cvpr}, ROVIO \cite{Bloesch15iros}).

The mapping operation is automatically started after an initial wait period (typically 0.5s) using poses from any of the above bootstrapping methods. 
Once an initial map is available, tracking also starts automatically. 
From this point onwards, the estimated camera poses produced by our event-based tracker are used for subsequent mapping.

\section{Experiments}
\label{sec:experim}

\subsection{Datasets and Metrics}
\input{floats/fig_dsec}
We evaluate the performance of our stereo VO pipeline on sequences from \emph{five} publicly available datasets \cite{Zhou18eccv,Zhu18ral,Gehrig21ral,Klenk21iros,Burner22evimo2} with varying camera resolutions depicting a wide range of scenarios on different mobile platforms.
Most of the datasets contain ground truth (GT) poses from a motion capture (mocap) system. 
Where camera poses are not available, like in DSEC \cite{Gehrig21ral}, we use LiDAR-inertial odometry data provided by the authors as ground truth. 
We use ground truth poses for accuracy assessment and also for bootstrapping.

Qualitative evaluation of our VO pipeline is done by visualizing the estimated camera trajectories and 6-DOF (degrees of freedom) pose, and comparing it to~ground truth (using \cite{Grupp17evo}), along with depth maps and point clouds.

Similar to recent event-based VO methods in the literature \cite{Klenk24threedv,Niu24icra,Zhou20tro},
we perform quantitative evaluation by computing the root-mean-square (RMSE) Absolute Trajectory Errors (ATE) and Absolute Rotation Errors (ARE) on tracked camera poses using the tool in \cite{Zhang18iros} (\cref{tab:metrics}). 
These combine the effects of both mapping and tracking estimation errors into a single metric.
This is because map errors leak into trajectory errors and vice-versa cumulatively, and trajectory errors are easier to assess than mapping errors.
Additionally, we carry out a broad evaluation of the performance of the improved mapper (\cref{fig:rayimprovements}) using ten standard quantitative metrics in \cref{sec:depth:ablation}.

In this work, we compare the camera tracking performance of our system with the state-of-the-art direct event-only stereo visual odometry method ESVO \cite{Zhou20tro} and with monocular system EVO \cite{Rebecq17ral}.
\Cref{tab:metrics} provides a comprehensive comparison of these methods on standard datasets of different resolutions.

It was not possible to compare to other methods in \cref{tab:relworks} because code was unavailable \cite{Hadviger21ecmr,Elmoudni23itsc,Wang23ral}
and/or trajectory errors were either not reported \cite{Elmoudni23itsc}, 
or did not use the same standard metrics as in ESVO (e.g., absolute errors) \cite{Hadviger21ecmr}.

Since our system is intended for live operation, experiments were performed in an online manner by playing ROSBags (slowed down for high resolution cameras), launching our tracking and mapping nodes, and recording the output poses and point clouds in real-time. 
Results may vary slightly depending on the CPU load, ROSBag playback speed and available processing power (as in ESVO).

\subsection{Results}
\subsubsection{RPG Stereo Dataset.}
This dataset was recorded with stereo DAVIS cameras with 240$\times$180 px resolution mounted on a handheld rig \cite{Zhou18eccv}. 
It comprises various indoor scenes inside a mocap room with 6-DOF camera motion. 
\Cref{tab:metrics} lists ATE and ARE for camera poses estimated by EVO, ESVO and our system on some commonly used sequences of this dataset.
For this dataset, we found different ATE values reported for ESVO in different papers, which may be due to its non-deterministic nature. 
Here, we report the values from the latest work by the same authors \cite{Niu24icra}. 
Since ARE is missing from that paper, we take them from \cite{Klenk24threedv}.
We observe that our system performed best, while EVO failed after completing at most 30\% of the \emph{bin} and \emph{boxes} sequences.

\input{floats/fig_mapping_snaps_dsec}
\subsubsection{DSEC Driving Dataset.}
The sequences in the DSEC dataset \cite{Gehrig21ral} were recorded with VGA resolution (640$\times$480 px) stereo event cameras on a car driven through the streets of Switzerland. 
Driving scenarios are challenging for event-based sensors because forward motions typically produce considerably fewer events in the center of the image (where apparent motion is small) than in the periphery. 
Forward motion is also particularly challenging for 3D reconstruction compared to sideways motions due to lack of motion parallax. 
For the evaluation, we used the \emph{zurich\_city\_04} sequences that contain many flashing lights (from vehicles and street signs) and Independently Moving Objects (IMOs), which makes it challenging for event-based motion estimation.

\Cref{fig:dsec_traj} shows small drift of the estimated trajectory of the car-mounted camera on all parts of the $\approx$2 km drive, 
whereas \cref{tab:metrics} reports that, compared to ESVO, our system is more accurate (in some cases by an order of magnitude). 
This is mainly due to better depth estimation performance of the ray denisty fusion-based mapping module compared to ESVO, as demonstrated in \cite{Ghosh22aisy, Elmoudni23itsc}, and further illustrated in \cref{fig:dsecdepthmaps} where we are able to reconstruct denser and sharper 3D edge maps that preserve details, even without map propagation.

The first 6s of the \emph{zurich\_city\_04\_c} sequence comprises a tram moving very close to the car and occupying majority of the cameras' FOV. 
The scene is dominated by events generated from independent motion, and our assumption of a static world becomes invalid, causing tracking failure. 
We thus report results after 6s, when the tram has moved sufficiently far away. 
It should be noted that the ESVO tracker also fails during this phase but thanks to its re-initialization module, it can recover and complete the sequence.

\subsubsection{TUM-VIE Dataset.} 
This dataset was recorded with stereo HD (1 Megapixel) event cameras mounted on a helmet and recorded from an egocentric viewpoint \cite{Klenk21iros}. 
We report results on sequences recorded in a mocap room containing varying DOFs of camera motion, spanning two distinct sets of calibration parameters. 
We also re-ran ESVO in this case.

A major issue in the mocap sequences is the large amount of noisy events generated by flashing (IR or LED) lights throughout the sequence. 
Particularly, the first few seconds of each sequence contain flashing fluorescent lights without any camera motion. 
We thus start both ESVO and our pipeline after the flashes subside (typically after 5s).
Moreover, the scene depth in these sequences is small relative to the camera baseline (\SI{11.84}{\cm}). 
If we set the reference view of our map on one camera trajectory, the large baseline makes the event rays back-projected from the other camera appear nearly parallel in the DSI, which hampers sharpening of edges during fusion.

Despite these challenges, \cref{tab:metrics} shows that we obtain better ATE in all but one sequence, and better ARE in 60\% of the evaluated sequences.
We also show qualitative results on the longest sequence \emph{desk}. 
\Cref{fig:tumvie-mapping} depicts the estimated global point cloud and camera trajectory for the full 33s sequence (with over 1 \emph{billion} events), along with snapshots of the events, the confidence map and the projected local point cloud. 
The camera performs multiple loops with 6-DOF motion in this sequence. 
Even without any back-end or map fusion through time, the resulting point cloud preserves details and sharp edges, showcasing minimal drift in our system.
\Cref{fig:tumvie} further compares the 6-DOF pose estimates over time against the ground truth and ESVO; the latter shows considerable drift in both translation and rotation.

Compared to EVO, our stereo VO pipeline produces more accurate results because it not only leverages temporal stereo, but also spatial parallax. 
This is especially noticeable in some of the sequences, like \emph{desk}, with an ATE improvement from 54.1 cm to 2.5 cm (\cref{tab:metrics}). 
Such improvement is also observed for the RPG dataset (see the top part of \cref{tab:metrics}).

\input{floats/fig_eyecatcher}

\subsubsection{EVIMO2 Trinocular Dataset.}
Adding more than two cameras to a sensor setup can improve 3D reconstruction by better handling occlusions and removing ambiguities \cite{Carneiro13nn}.
Most stereo methods estimate depth by matching features across image pairs. 
For $N$ cameras, this involves a combinatorial number of matching steps, 
${}^N\!C_2$ times (3 times for 3 cameras, 12 times for 4 cameras, etc.). 
This makes multi-camera stereo prohibitively expensive for real-time SLAM/VO despite the promised benefits.
By foregoing the need for finding pairwise correspondences, our mapping module scales with linear complexity -- for $N$ cameras, the DSI creation and fusion steps are done $N$ times (see \cite{Ghosh22aisy}). 
Our system thus unlocks the use of multi-camera setups for event-based VO/SLAM for the first time.

We present results on the only existing trinocular event dataset EVIMO2 \cite{Burner22evimo2}, recorded using a handheld rig with three VGA resolution event cameras in an indoor environment. 
Unfortunately, there is narrow overlap between the FOVs of the cameras because the Prophesee Gen3 cameras on either side are in portrait mode, whereas the Samsung DVS, in the center, is in landscape mode. 
This makes depth estimation difficult when the cameras approach too close to the objects placed on a table, as they may fall outside the overlapping FOV.
We set the central camera as the reference for both tracking and mapping.

\Cref{fig:evimo2} shows the camera trajectory and global semi-dense point cloud estimated by our VO system as it goes around a table top.
The system exhibits very good tracking accuracy when there is sufficient overlap among the camera views.
For comparison, we also tried a stereo configuration with just the left and right event cameras, but the VO system lost track easily due to the smaller overlap between the cameras (compared to the trinocular configuration). 
This experiment exemplifies the benefit of having redundancy in the trinocular system.

\input{floats/fig_mvsec}

\subsubsection{MVSEC Dataset.}
Finally, we also present results on the MVSEC dataset \cite{Zhu18ral}.
The low spatial resolution (346$\times$260 px), small baseline of the stereo rig and few events generated far from the camera are challenging factors for accurate pose estimation. 
The outdoor sequences are not usable in stereo since the baseline is too small \cite{Ghosh22aisy} compared to a 1-pixel disparity resolution.

We thus show qualitative results on the \emph{indoor\_flying1} sequence, which comprises arbitrary 6-DOF motion from a camera-mounted drone flying inside a mocap room. 
The quick forward and backward motion of the drone combined with hovering halts (with almost no events generated) in between make it especially challenging. 
The drone also often comes very close to the reflective and salt-and-pepper textured ground surface, 
which generates many unstructured brightness changes, further challenging event-based VO.

Nevertheless, our system successfully tracks long stretches of the flight involving to-fro motion, sudden change of direction, hoverings and loops, as shown in \cref{fig:mvsec}, which depicts the estimated trajectory and global point cloud, and the individual camera DoFs compared with ground truth. 
While our system estimates translations accurately, we observe small accumulation of drift in the orientation.

\subsection{Depth Estimation Ablation}
\label{sec:depth:ablation} 
The results of depth estimation of our mapper on the full DSEC \emph{zurich04a} driving sequence are reported in \cref{tab:dsec}. 
We followed the evaluation protocol in \cite{Ghosh22aisy}, as detailed in the caption. 
For MC-EMVS and EMVS, the DSI reference views were set at the camera trajectory center.
The last row corresponds to our improved MC-EMVS (\cref{sec:method:mapper}), where the reprojected rays are limited to a minimum depth plane, making it robust to all forward and backward motions. 
With this change, camera centers are not mistaken for 3D structures (see \cref{fig:rayimprovements}), significantly improving depth estimation accuracy (32.5\% drop in log-RMSE).

\input{floats/tab_mapping}

\subsection{Computational Performance}
Tests were run on a desktop computer with CPUs of Intel Xeon(R) W-2225 at 4.10GHz$\times$8 cores.
The runtime numbers for processing MVSEC data (stereo DAVIS346 cameras with 346$\times$260 px resolution) are reported in \cref{tab:runtime}.
As observed, the slowest component of the system is the DSI creation in the mapper. 
This processing time can be reduced by decreasing the number of depth planes (at the expense of some loss of precision in depth) or by parallelizing using GPUs.
Nevertheless, this is not always needed: in the accompanying video, we demonstrate live operation of our system with a stereo DAVIS346 setup.

\input{floats/tab_runtime}

\subsection{Limitations}

While the parallel-tracking-and-mapping strategy works and is common to many approaches, 
the propagation of errors, and therefore the growth of drift is inevitable. 
This could be mitigated by incorporating a refinement step (bundle adjustment) in the system.
This is an emerging avenue for future research \cite{Guo24tro}.

Furthermore, most SLAM methods lack a strategy to exclude such large distractors (IMOs), which can cause drift (as observed in the experiments). 
Dealing with them, possibly via a combination with motion segmentation \cite{Zhou21tnnls}, is still an unexplored topic in the incipient field of event-based SLAM.

\section{Conclusion}
\label{sec:conclusion}

We have introduced a novel event-only stereo visual odometry system.
It has a geometric design, combining a mapping module based on the idea of ray density fusion, with improved capabilities over MC-EMVS, 
and a tracking module based on the idea of global edge-map alignment.
The proposed method has been thoroughly tested on five datasets, more than any of the previous works, and it outperforms the state-of-the-art ESVO and EVO methods by a margin.
The experiments show the remarkable capabilities of our method to handle a wide range of scenes, event camera types and spatial resolutions (from HQVGA to HD), outputting accurate camera trajectories and sharp semi-dense maps.
Future exciting research directions for improvement comprise large IMO handling and a back-end refinement.
We release the source code and 
hope that our work fosters research into event-based VO methods for autonomous vehicles and robots.

\section*{Acknowledgments}
Funded by the SONY Research Award Program 2021.
Funded by the Deutsche Forschungsgemeinschaft (DFG, German Research Foundation) under Germany’s Excellence Strategy -- EXC 2002/1 ``Science of Intelligence'' -- project number 390523135.

\bibliographystyle{splncs04}
%\bibliography{all}

\input{main.bbl}
\end{document}

%% file: macros.tex
\def\Lum{L}
\def\pol{p} % event polarity
\def\cE{\mathcal{E}} % Events
\def\numEvents{N_e} % number of events
\def\Warp{\mathbf{W}}

\def\bx{\mathbf{x}}

\def\bp{\mathbf{p}}

\def\mP{\mathtt{P}}

\def\bX{\mathbf{X}}

\robustify \bfseries
\robustify \underline
\definecolor{light-gray}{gray}{0.75}
\newcommand\gframe[1]{\color{light-gray}\frame{#1}}

%% file: floats/fig_mapping_tumvie.tex
% Two figures in one
\begin{figure}[t]
    \centering
    % -------------------------------------------------------------------------
    \begin{subfigure}[c]{0.49\linewidth}
         \centering
\def\figWidth{0.315\linewidth}         
{\scriptsize
    \setlength{\tabcolsep}{2pt}
	\begin{tabular}{
	>{\centering\arraybackslash}m{\figWidth}
	>{\centering\arraybackslash}m{\figWidth}
	>{\centering\arraybackslash}m{\figWidth}}

          % \gframe
        \multicolumn{3}{c}{ %\gframe
        {\includegraphics[trim={1cm 0.5cm 0 5cm},clip,width=.97\linewidth]{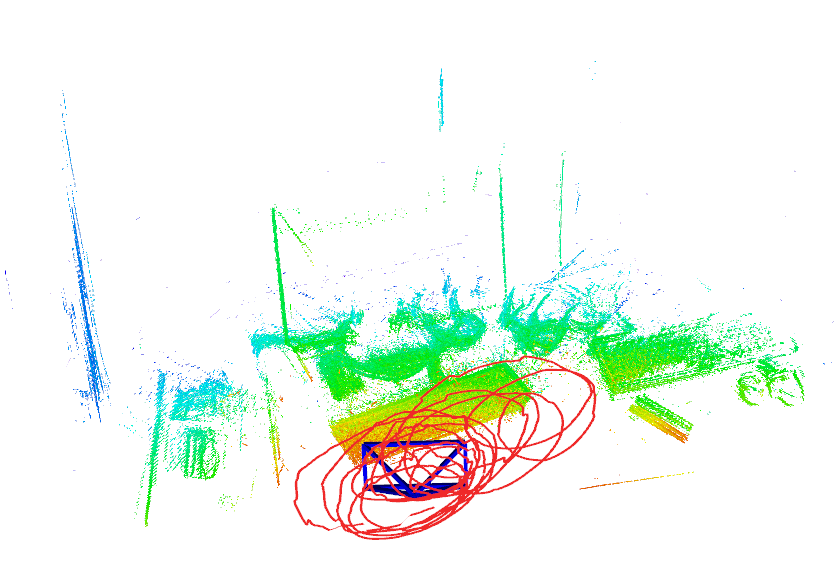}}}\\
        
	     Events (left camera) & Confidence map & Projected map on events\\
      
		% \gframe{\includegraphics[trim={0px 130px 0 200px},clip,width=\linewidth]{images/tumvie/bike-easy-3/01558.jpg}}
		% &\gframe{\includegraphics[trim={0px 0 150px 0},clip,width=\linewidth]{images/tumvie/bike-easy-3/hdr_events_bike.png}}
  %       &\gframe{\includegraphics[trim={0px 0 200px 0},clip,width=\linewidth]{images/tumvie/bike-easy-3/14.000000inv_depth_colored_dilated_fused_2_w.png}}
		% &\gframe{\includegraphics[trim={0px 0 150px 0},clip,width=\linewidth]{images/tumvie/bike-easy-3/14.000000inv_depth_colored_dilated_fused_2_w.png}}
		% \\
		
		% \gframe{\includegraphics[trim={0px 150px 0 300px},clip,width=\linewidth]{images/tumvie/skate-easy-1/00686.jpg}}
		% &\gframe{\includegraphics[trim={0px 0 0 0},clip,width=\linewidth]{images/tumvie/skate-easy-1/hdr_events_skate.png}}
		% &\gframe{\includegraphics[trim={0px 0 0 0},clip,width=\linewidth]{images/tumvie/skate-easy-1/15.000000inv_depth_colored_dilated_fused_2_w.png}}
  %       &\gframe{\includegraphics[trim={0px 0 0 0},clip,width=\linewidth]{images/tumvie/skate-easy-1/15.000000inv_depth_colored_dilated_fused_2_w.png}}
		% \\
  
         \gframe{\includegraphics[trim={0px 0 0 0},clip,width=\linewidth]{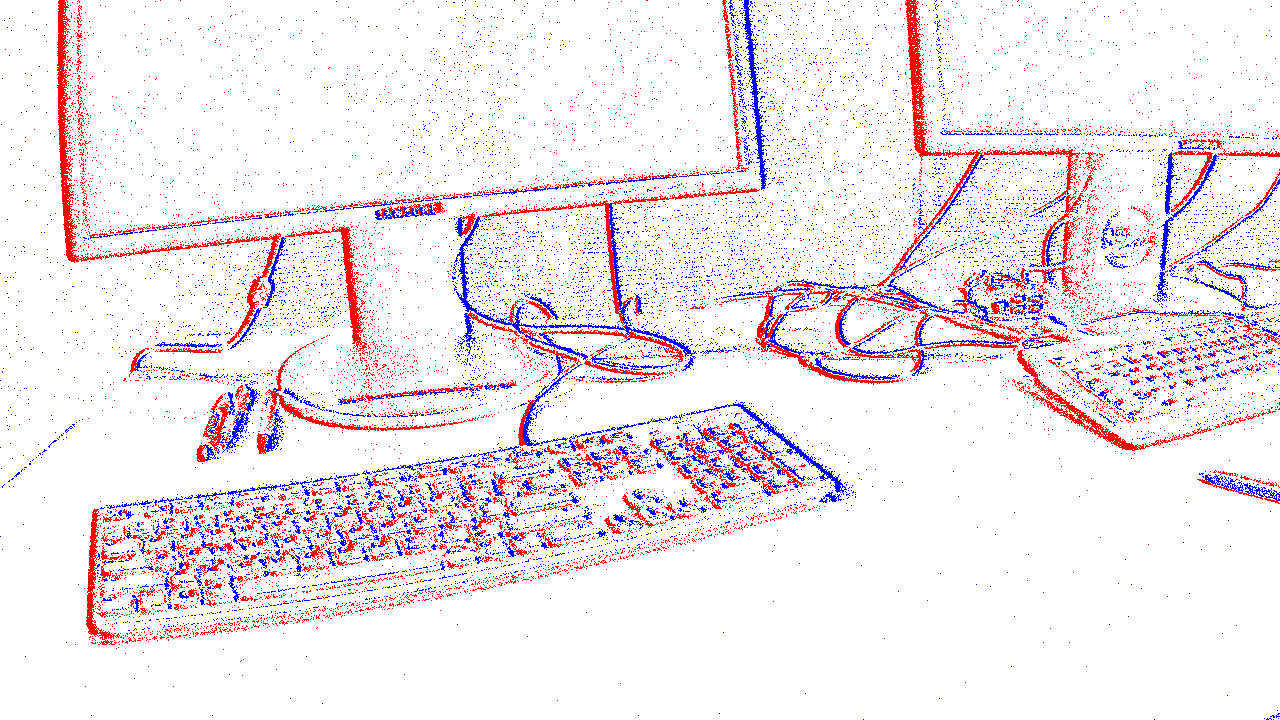}}
		&\gframe{\includegraphics[trim={0px 0 0 0},clip,width=\linewidth]{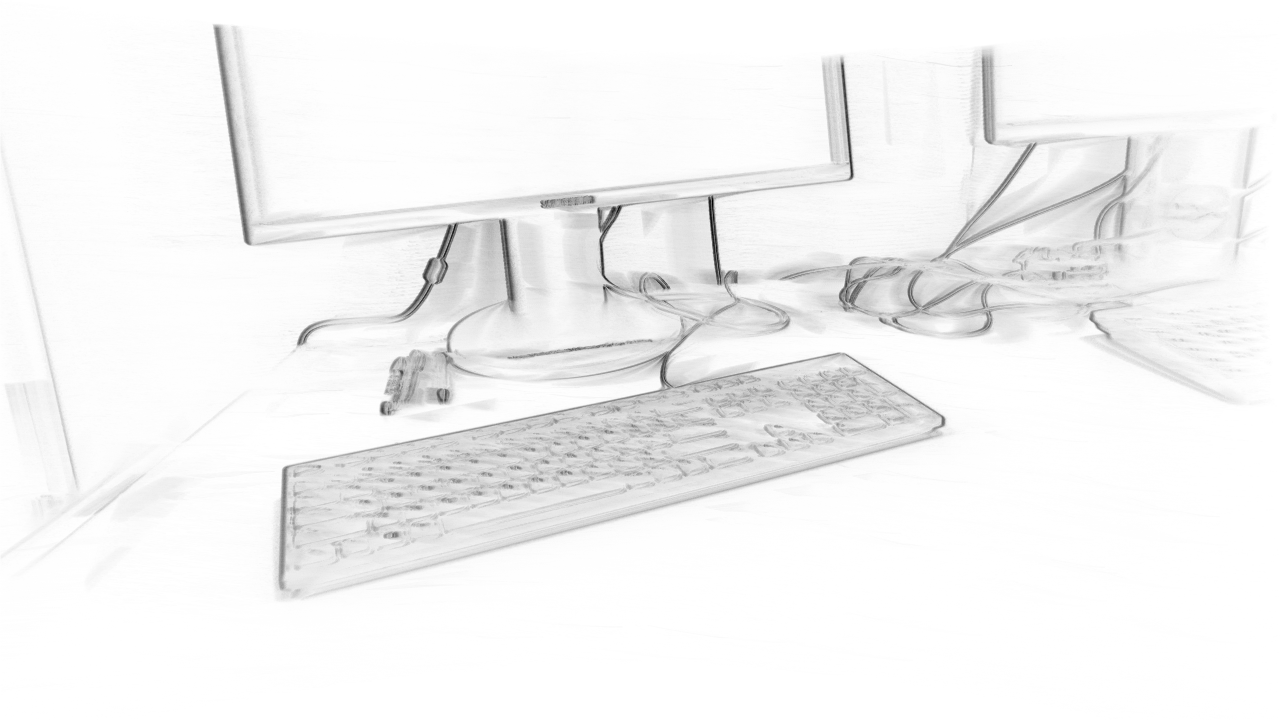}}
        &\gframe{\includegraphics[trim={0px 0 0 0},clip,width=\linewidth]{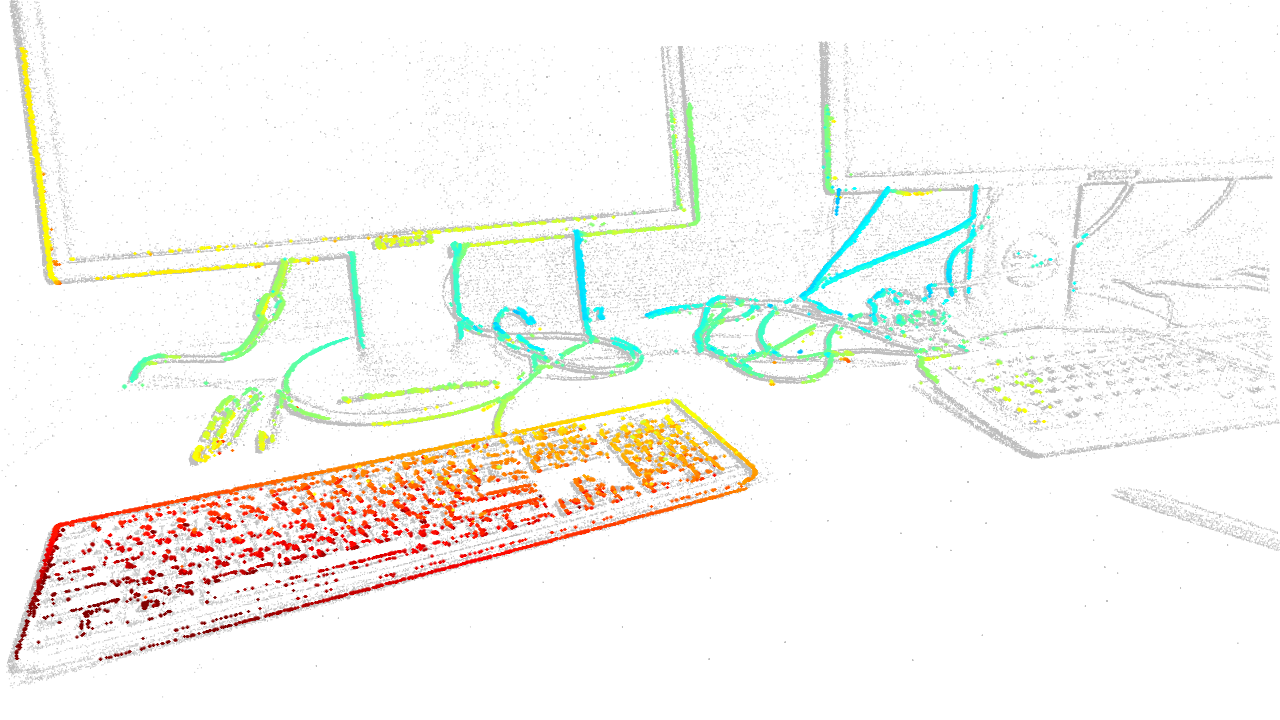}}
	\end{tabular}
	}
         
         \caption{Overview}
         \label{fig:eyecatcher}
    \end{subfigure}
    % -------------------------------------------------------------------------
    \begin{subfigure}[c]{0.49\linewidth}
         \centering
{\scriptsize
\begin{tabular}{ccc}
    Translation & \, & Rotation \\
    %\gframe
    {\includegraphics[trim={0cm 0.5cm 0cm 0cm},clip,width=.48\linewidth]{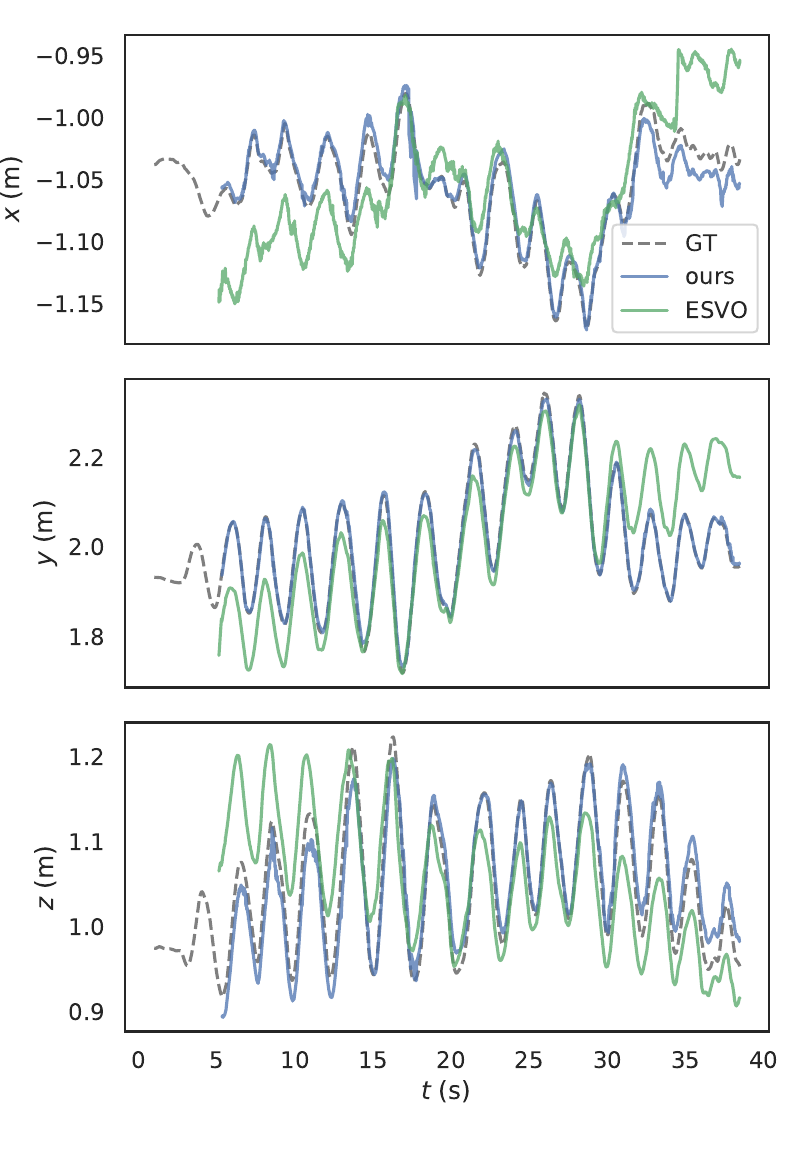}}
    & & 
    %\gframe
    {\includegraphics[trim={0cm 0.5cm 0cm 0cm},clip,width=.48\linewidth]{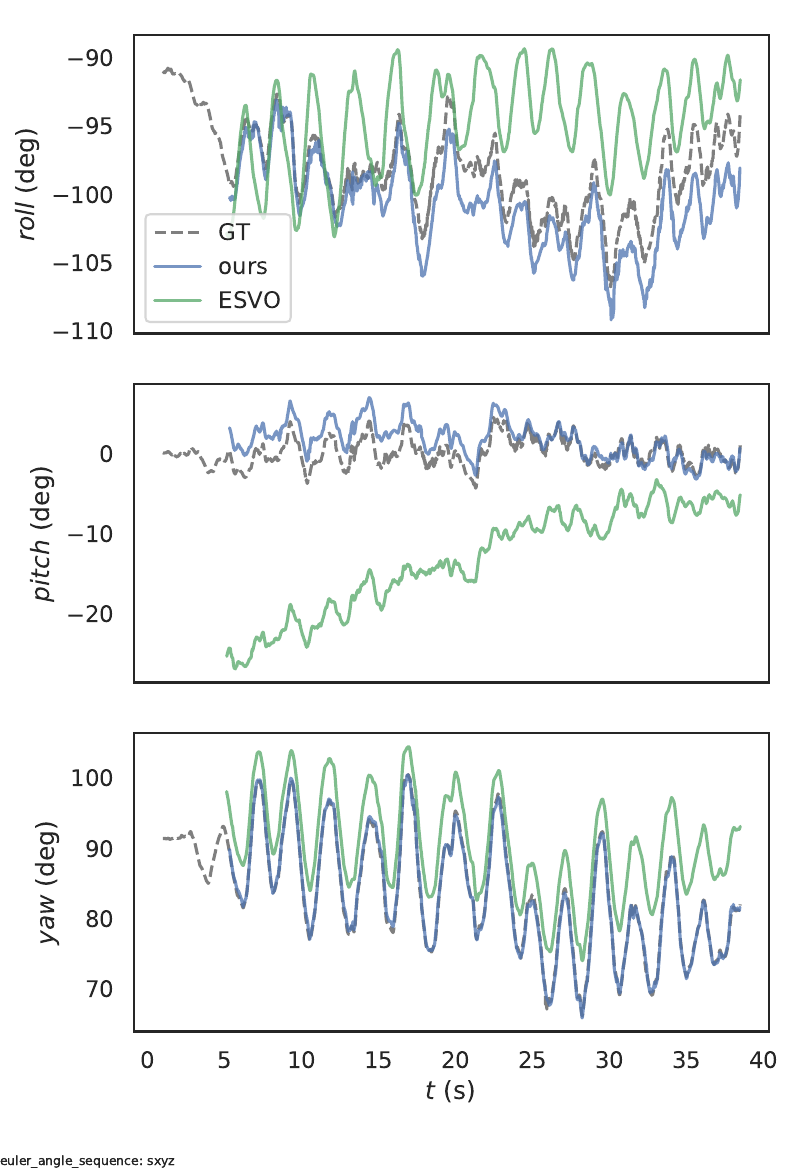}}
\end{tabular}
}
         \caption{Camera Trajectory}
         \label{fig:tumvie}
     \end{subfigure}
    % -------------------------------------------------------------------------
    \caption{(a) 3D point cloud and camera trajectory estimated by our stereo VO pipeline for the TUM-VIE \emph{mocap-desk} sequence \cite{Klenk21iros}, along with a snapshot of events, confidence map and the projected point cloud (overlaid on the events). 
    Depth is color-coded from red (near) to blue (far away).
    (b) Estimated camera poses over time, compared to ground truth (GT) and ESVO \cite{Zhou20tro}.
    }
    \label{fig:tumvie-mapping}
\end{figure}

%% file: floats/tab_related_works.tex
\newcommand{\cmark}{\ding{51}}%
\newcommand{\xmark}{\ding{55}}%

\begin{table*}[t]
\centering
\caption{Stereo event-only VO methods for 6-DOF ego-motion estimation, sorted chronologically.
    The column D/I indicates whether the method is \textbf{D}irect or \textbf{I}ndirect (feature-based);
    the column ``Map Prop.'' indicates whether local maps are propagated in time and fused or are local only.}
    \label{tab:relworks}
    \begin{adjustbox}{max width=\textwidth}
    \setlength{\tabcolsep}{4pt}
    \begin{tabular}{@{}lclllcl@{}}
    \toprule
    \textbf{Method}          & \textbf{D/I} & \textbf{Event representation} & \textbf{Tracking objective} & \textbf{Mapping objective} & \textbf{Map Prop.} & \textbf{Datasets evaluated on} \\
    \midrule
    
    ESVO \cite{Zhou20tro}  & D & Time surfaces (TS) & \makecell[tl]{TS as anisotropic\\ distance field} 
    & \makecell[tl]{Stereo matching\\TS patches} & \cmark & RPG, MVSEC, HKUST \\
    
    Hadviger \cite{Hadviger21ecmr} & I & Arc$^\ast$ corners on TS & Min. reproj. error & \makecell[tl]{Cross-corr. feature\\ descriptors} & \xmark & MVSEC, DSEC \\
    
    Wang \cite{Wang23ral} & I & Libviso2 on binary %event 
    images %, SAE 
    & Min. reproj. error & Feature matching & \xmark & MVSEC \\[1em]
    
    El Moudni \cite{Elmoudni23itsc}  & D & Raw \& Time surfaces (TS) & \makecell[tl]{TS as anisotropic\\ distance field} & Ray density fusion & \xmark & DSEC \\
    
    Shiba \cite{Shiba24pami} & D & Raw & Max. Contrast & Max. Contrast & \xmark & DSEC \\[1em]
    
    \textbf{This work} & D & Raw \& binary event images & \makecell[tl]{Photometric edge\\ alignment} & Ray density fusion & \xmark & \makecell[tl]{RPG, MVSEC, DSEC\\ TUM-VIE, EVIMO2} \\
    \bottomrule
    \end{tabular}
    \end{adjustbox}
\end{table*}

%% file: floats/fig_pipeline.tex
\begin{figure}[t]
    \centering
    {\includegraphics[trim={0.6cm 0.35cm 0.6cm 0.34cm},clip,width=\linewidth]{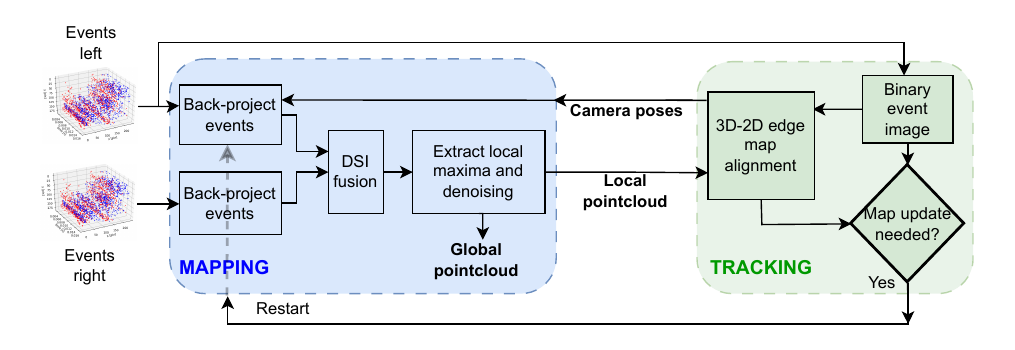}}
    \caption{Proposed event-based stereo visual odometry pipeline consisting of two main modules: camera tracking (i.e., ego-motion estimation) and scene mapping.}
    \label{fig:pipeline}
\end{figure}

%% file: floats/fig_ray_improvements.tex
% \def\figWidth{0.38\linewidth}
% \begin{figure}[t]
% 	\centering
%     {\small
%     \setlength{\tabcolsep}{2pt}
% 	\begin{tabular}{
%     >{\centering\arraybackslash}m{0.3cm} 
% 	>{\centering\arraybackslash}m{\figWidth}
% 	>{\centering\arraybackslash}m{\figWidth}}
% 	    & MC-EMVS \cite{Ghosh22aisy} & Ours \\

%         \rotatebox{90}{\makecell{Confidence map}}
%         &\gframe{\includegraphics[trim={0px 0 0 0},clip,width=\linewidth]{images/new_ray_casting/confidence_before_ray_limit_gamma.png}}
%         &\gframe{\includegraphics[trim={0px 0 0 0},clip,width=\linewidth]{images/new_ray_casting/confidence_after_ray_limit_gamma.png}}
% 		\\

%         \rotatebox{90}{\makecell{Depth map}}
%         &\gframe{\includegraphics[trim={0px 0 0 0},clip,width=\linewidth]{images/new_ray_casting/depth_before_ray_limit_box.png}}
%         &\gframe{\includegraphics[trim={0px 0 0 0},clip,width=\linewidth]{images/new_ray_casting/depth_after_ray_limit_box.png}}
% 		\\
  
% 	\end{tabular}
% 	}
% 	\caption{Our improved ray casting produces clean confidence and depth maps. 
%     See its effect near the focus of expansion.
%     Depth is color-coded from red (near) to blue (far away).
% 	}
%     \label{fig:rayimprovements}
% \end{figure}

\def\figWidth{0.238\linewidth}
\begin{figure}[t]
	\centering
    {\small
    \setlength{\tabcolsep}{2pt}
	\begin{tabular}{
	>{\centering\arraybackslash}m{\figWidth}
    >{\centering\arraybackslash}m{\figWidth}
    >{\centering\arraybackslash}m{0.01cm} 
    >{\centering\arraybackslash}m{\figWidth}
	>{\centering\arraybackslash}m{\figWidth}}
	    \multicolumn{2}{c}{MC-EMVS \cite{Ghosh22aisy}} 
        & 
        & \multicolumn{2}{c}{Ours} \\

        %\rotatebox{90}{\makecell{Confidence map}}
        \gframe{\includegraphics[trim={0px 0 0 0},clip,width=\linewidth]{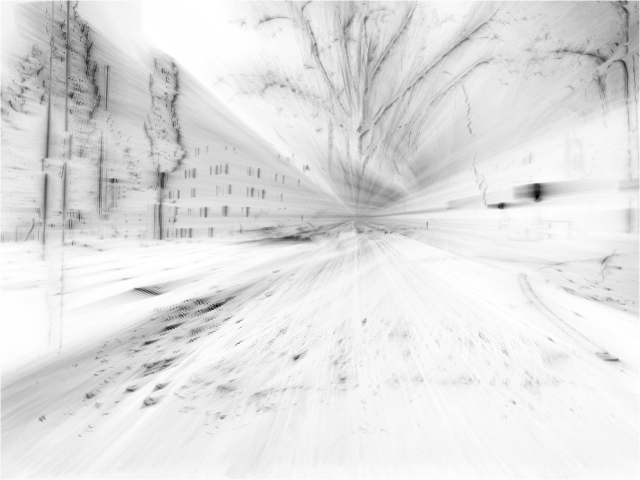}}
        &\gframe{\includegraphics[trim={0px 0 0 0},clip,width=\linewidth]{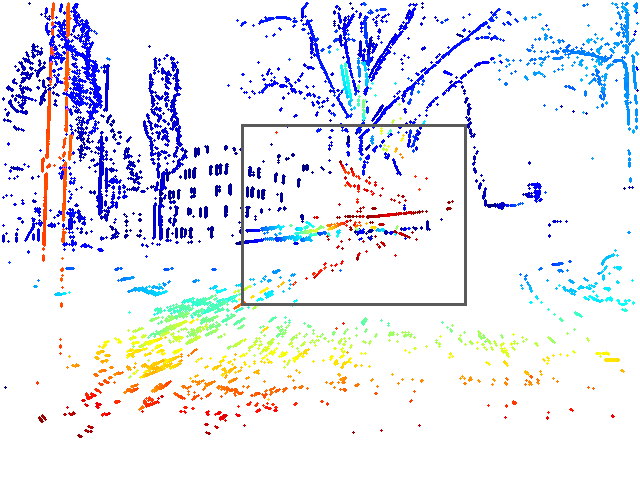}}
        &
        %\rotatebox{90}{\makecell{Depth map}}
        &\gframe{\includegraphics[trim={0px 0 0 0},clip,width=\linewidth]{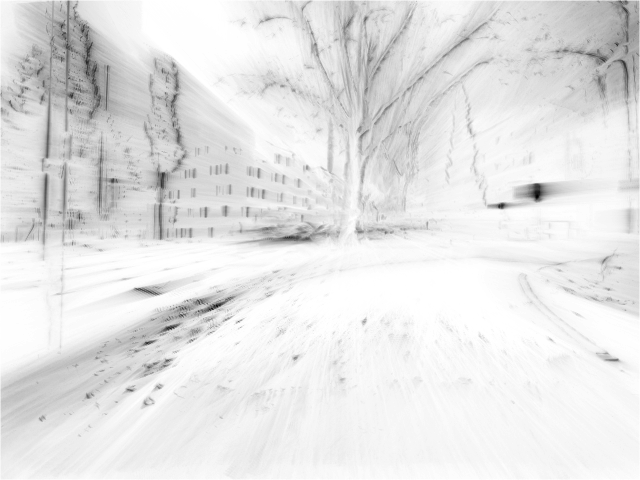}}
        &\gframe{\includegraphics[trim={0px 0 0 0},clip,width=\linewidth]{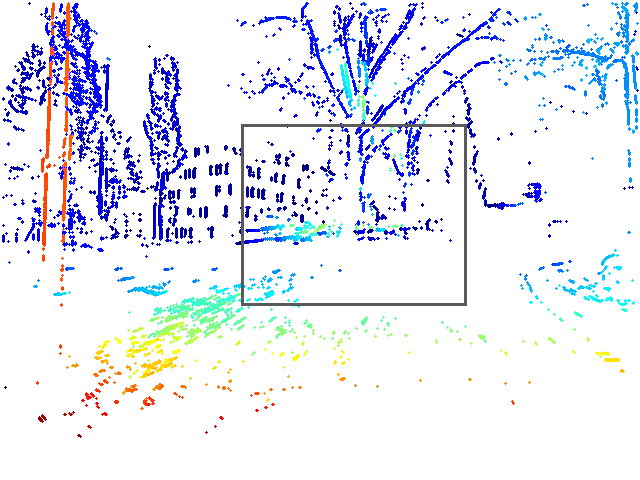}}
		\\
        
        Confidence map
        & Depth map
        & 
        & Confidence map
        & Depth map\\  
	\end{tabular}
	}
	\caption{Our improved ray casting produces clean confidence and depth maps. 
    See its effect near the focus of expansion.
    Depth is color-coded from red (near) to blue (far).
	}
    \label{fig:rayimprovements}
\end{figure}

%% file: floats/tab_metrics_all.tex
\begin{table*}[t]
\centering
\caption{Absolute Trajectory Error (ATE) and Absolute Rotation Error \cite{Zhang18iros} (ARE) on datasets of multiple resolutions (and therefore varying event rates).
All the values for EVO \cite{Rebecq17ral} are taken from \cite{Klenk24threedv}. 
ESVO values are taken from the authors' latest paper \cite{Niu24icra}, except for the TUM-VIE dataset on which we re-ran ESVO \cite{Zhou20tro}. 
Rotation errors on the RPG dataset were taken from \cite{Klenk24threedv}.
(-) indicates missing values. 
($^\ast$) indicates failure in after completing at most 30\% of the sequence.
($^\dagger$) indicates when the first 6s were skipped because of large independent motion in the system's FOV.
}
\label{tab:metrics}
\begin{adjustbox}{max width=\linewidth}
\setlength{\tabcolsep}{5pt}
\begin{tabular}{llrrr|rrr|rrr}\toprule
\textbf{Dataset} &\textbf{Seq.} &\textbf{Time} &\textbf{Length} &\textbf{ \#Ev./cam} &\multicolumn{3}{c}{\textbf{ATE RMSE [cm]}} & \multicolumn{3}{c}{\textbf{ARE RMSE [deg]}}\\[0.5ex]
& &\textbf{[s]} &\textbf{[cm]} & \textbf{[million]} &\textbf{EVO} &\textbf{ESVO} &\textbf{Ours} &\textbf{EVO} &\textbf{ESVO} &\textbf{Ours} \\

\midrule
RPG \cite{Zhou18eccv}
&monitor & 23 & 675 & 13 & 7.8 &5.8 &\textbf{2.34} & 7.77 & 2.74 & \textbf{1.52} \\
(240$\times$180 px)
&desk & 13 & 403 & 9 & 5.2 &7.25 &\textbf{2.84} & 8.25 & 7.25 & \textbf{3.44}\\
&bin & 17 & 514 & 9 & 13.2$^\ast$ & - &\textbf{2.57} & 50.26$^\ast$ & 7.61 & \textbf{3.29} \\
&boxes & 15 & 704 & 12 & 14.2$^\ast$ &9.5 &\textbf{4.06} & 170.36$^\ast$ & 9.46 & \textbf{6.62} \\

\midrule
DSEC \cite{Gehrig21ral} 
&zc04a & 35.0 & 23172 & 359 &-  &371.1 &\textbf{131.62} &- &- & \textbf{3.17} \\
(640$\times$480 px)
&zc04b & 13.4 & 5888 & 130 &-  &116.6 &\textbf{29.02}  &- &- & \textbf{2.04} \\
&zc04c & 59.0 & 52465 & 391 &-  &1357.1 &\textbf{1184.37$^\dagger$}  &- &- & \textbf{6.02}$^\dagger$\\
&zc04d & 47.8 & 53994 & 325 &-  &2676.6 &\textbf{1053.87}  &- &- & \textbf{37.13}\\
&zc04e & 13.6 & 12278 & 119 &-  &794.9 &\textbf{75.9}  &- &- & \textbf{3.97} \\
&zc04f & 43.1 & 38758 & 362 &- &- &\textbf{522}  &- &- & \textbf{10.65}\\

\midrule
TUM-VIE \cite{Klenk21iros}
&1d-trans & 36.6 & 502 & 573 & 7.5  & 12.31 &\textbf{1.05} &- & 7.77 & \textbf{6.02}\\
(1280$\times$720 px)
&3d-trans & 33.2 & 719 & 871 & 12.5  & 17.17 &\textbf{8.53} &- & 33.67 & \textbf{15.62}\\
&6dof & 19.5 & 531 & 568 & 85.5 & 13.04 &\textbf{10.25} &- & \textbf{10.76} & 14.01\\
&desk & 37.5 & 1033 & 1130 & 54.1 & 12.38 &\textbf{2.5} &- & 17.48 & \textbf{3.37}\\
&desk2 & 21.4 & 568 & 643 & 75.2 & \textbf{4.56} &7.2 &- & \textbf{5.09} & 10.12\\
\bottomrule
\end{tabular}
\end{adjustbox}
\end{table*}

%% file: floats/fig_dsec.tex
\begin{figure}[t]
\centering
    % \gframe
    {\includegraphics[trim={1.5cm 1cm 0.5cm 2cm},clip,width=.65\linewidth]{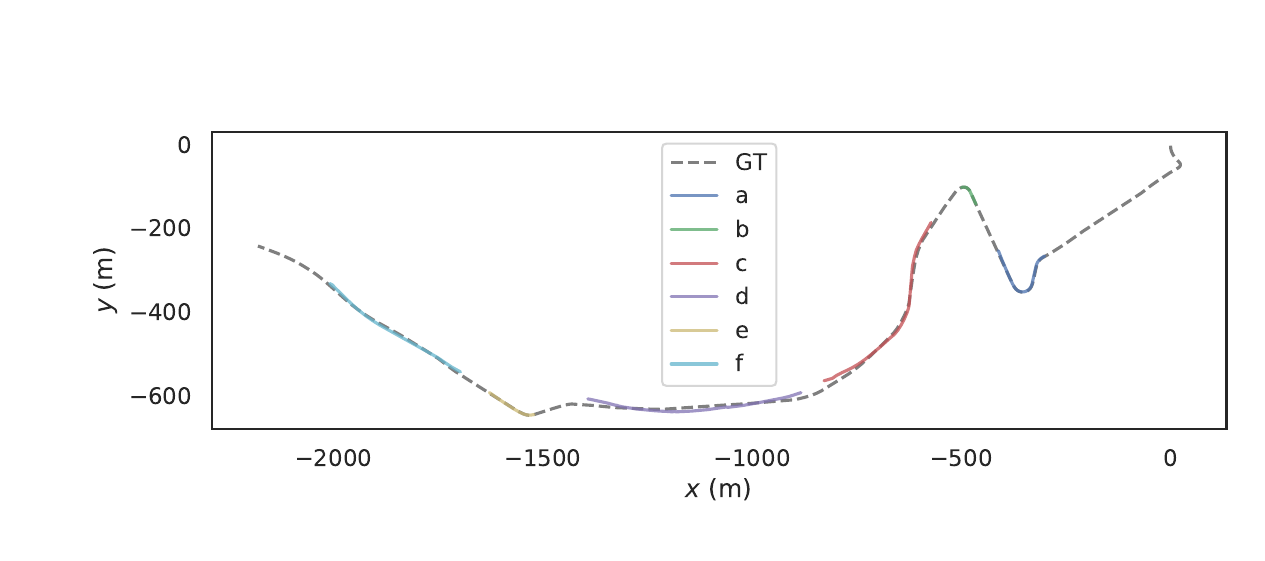}}
	\caption{Estimated trajectory of \emph{zurich\_city\_04} sequence from the DSEC dataset vs ground truth.
    The six evaluation segments (a)-(f) correspond to the rows of \cref{tab:metrics}.}
\label{fig:dsec_traj}
\end{figure}

%% file: floats/fig_mapping_snaps_dsec.tex
\def\figWidth{0.24\linewidth}
\begin{figure}[t]
	\centering
    {\scriptsize
    \setlength{\tabcolsep}{2pt}
	\begin{tabular}{
	>{\centering\arraybackslash}m{\figWidth}
	>{\centering\arraybackslash}m{\figWidth}
	>{\centering\arraybackslash}m{\figWidth}}
	     Events (left camera) & ESVO\cite{Zhou20tro} depth map& Our depth map\\
      
		% \gframe{\includegraphics[trim={0px 130px 0 200px},clip,width=\linewidth]{images/tumvie/bike-easy-3/01558.jpg}}
		% &\gframe{\includegraphics[trim={0px 0 150px 0},clip,width=\linewidth]{images/tumvie/bike-easy-3/hdr_events_bike.png}}
  %       &\gframe{\includegraphics[trim={0px 0 200px 0},clip,width=\linewidth]{images/tumvie/bike-easy-3/14.000000inv_depth_colored_dilated_fused_2_w.png}}
		% &\gframe{\includegraphics[trim={0px 0 150px 0},clip,width=\linewidth]{images/tumvie/bike-easy-3/14.000000inv_depth_colored_dilated_fused_2_w.png}}
		% \\
		
		% \gframe{\includegraphics[trim={0px 150px 0 300px},clip,width=\linewidth]{images/tumvie/skate-easy-1/00686.jpg}}
		% &\gframe{\includegraphics[trim={0px 0 0 0},clip,width=\linewidth]{images/tumvie/skate-easy-1/hdr_events_skate.png}}
		% &\gframe{\includegraphics[trim={0px 0 0 0},clip,width=\linewidth]{images/tumvie/skate-easy-1/15.000000inv_depth_colored_dilated_fused_2_w.png}}
  %       &\gframe{\includegraphics[trim={0px 0 0 0},clip,width=\linewidth]{images/tumvie/skate-easy-1/15.000000inv_depth_colored_dilated_fused_2_w.png}}
		% \\
  		
        \gframe{\includegraphics[trim={0px 0 0 0},clip,width=\linewidth]{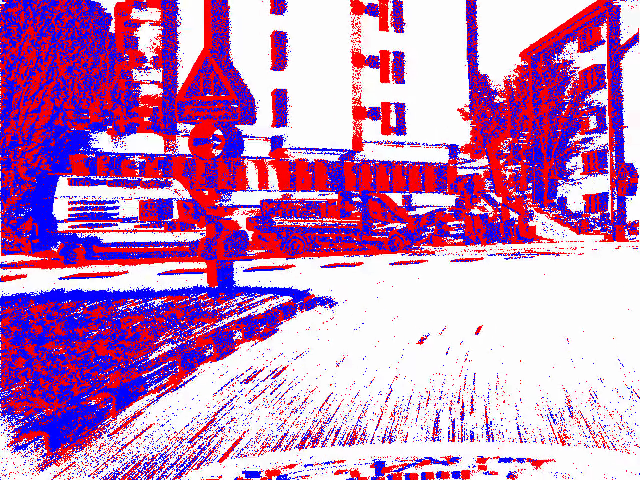}}
		&\gframe{\includegraphics[trim={0px 0 0 0},clip,width=\linewidth]{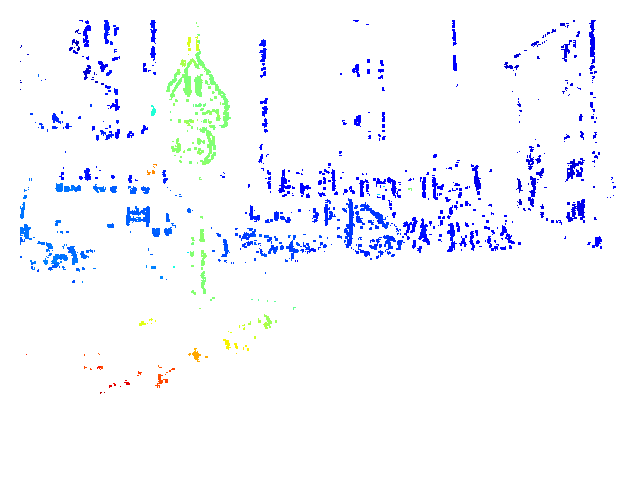}}
        &\gframe{\includegraphics[trim={0px 0 0 0},clip,width=\linewidth]{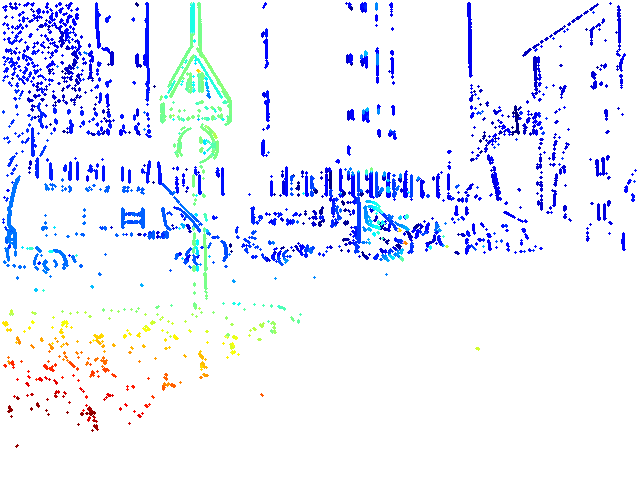}}
		\\
  
        \gframe{\includegraphics[trim={0px 0 0 0},clip,width=\linewidth]{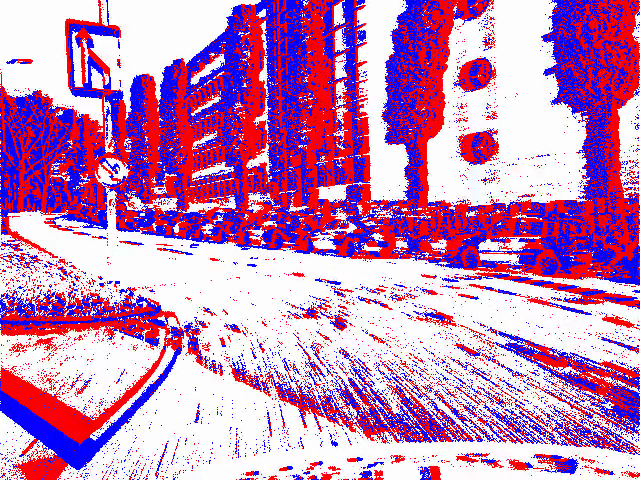}}
		&\gframe{\includegraphics[trim={0px 0 0 0},clip,width=\linewidth]{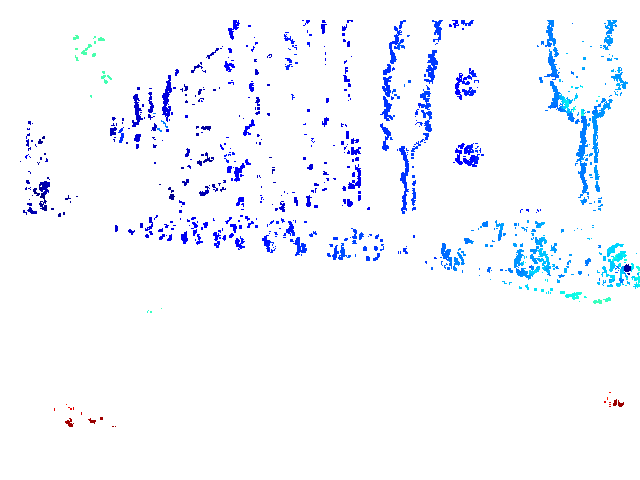}}
        &\gframe{\includegraphics[trim={0px 0 0 0},clip,width=\linewidth]{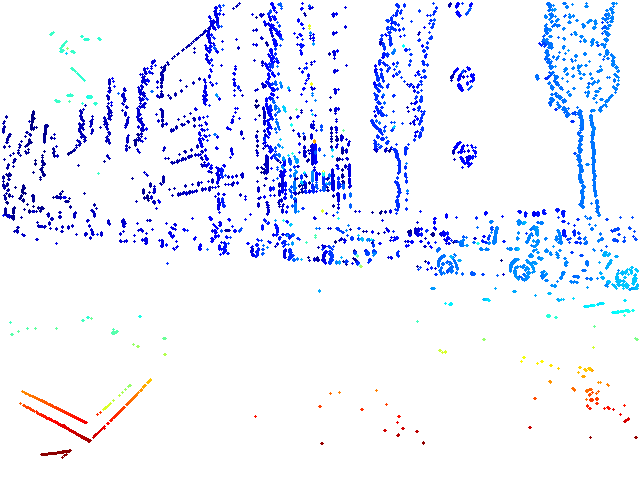}}
		\\

  		\gframe{\includegraphics[trim={0px 0 0 0},clip,width=\linewidth]{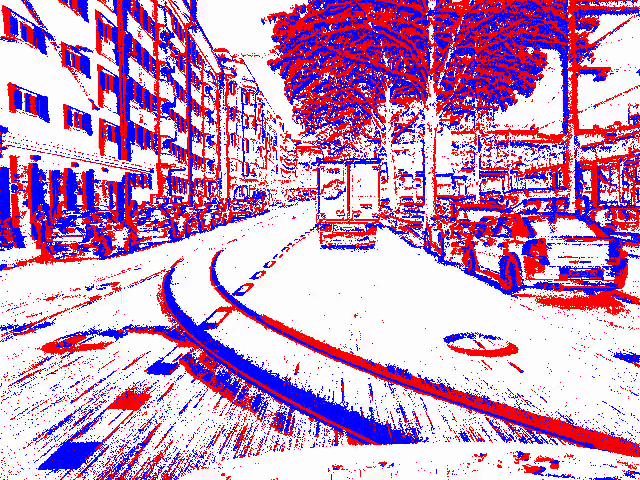}}
		&\gframe{\includegraphics[trim={0px 0 0 0},clip,width=\linewidth]{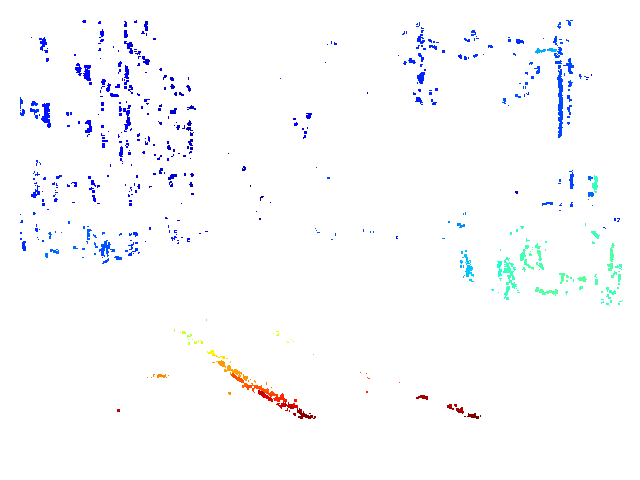}}
        &\gframe{\includegraphics[trim={0px 0 0 0},clip,width=\linewidth]{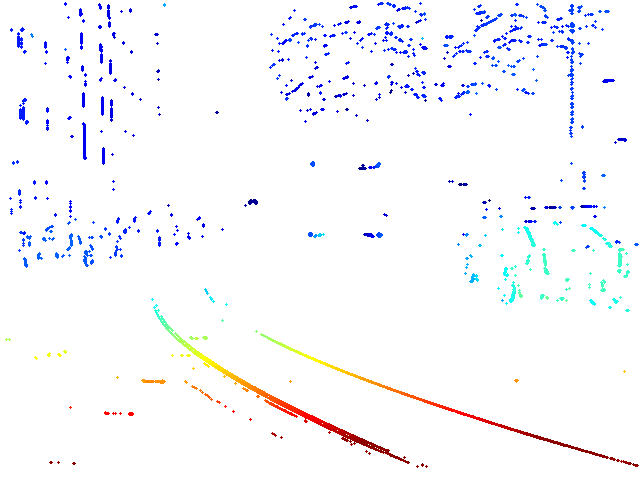}}
		\\
  
	\end{tabular}
	}
	\caption{Results on DSEC using our method and ESVO.
    The sharper edge maps from our method allow us to track camera poses accurately using edge alignment.}
    \label{fig:dsecdepthmaps}
\end{figure}

%% file: floats/fig_eyecatcher.tex
% Two figures in one
\begin{figure}[t]
    \centering
    \begin{subfigure}[c]{0.54\linewidth}
         \centering
         \includegraphics[trim={0 8cm 11.5cm 0},clip,width=\linewidth]{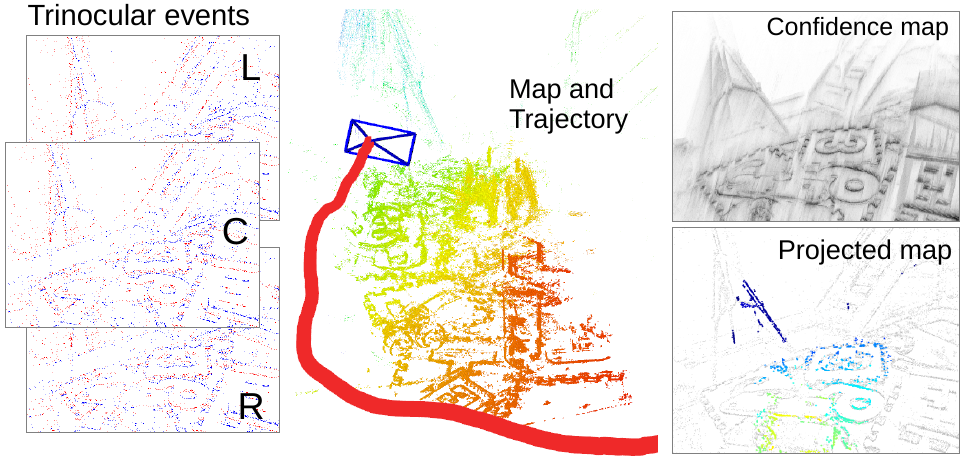}
         \caption{Overview}
         \label{fig:evimo2:rviz}
    \end{subfigure}
    \,
    \begin{subfigure}[c]{0.43\linewidth}
         \centering
{\small
\begin{tabular}{ccc}
    Translation & \, & Rotation \\
    % \gframe
    {\includegraphics[trim={0cm 0.5cm 0cm 0.3cm},clip,width=.45\linewidth]{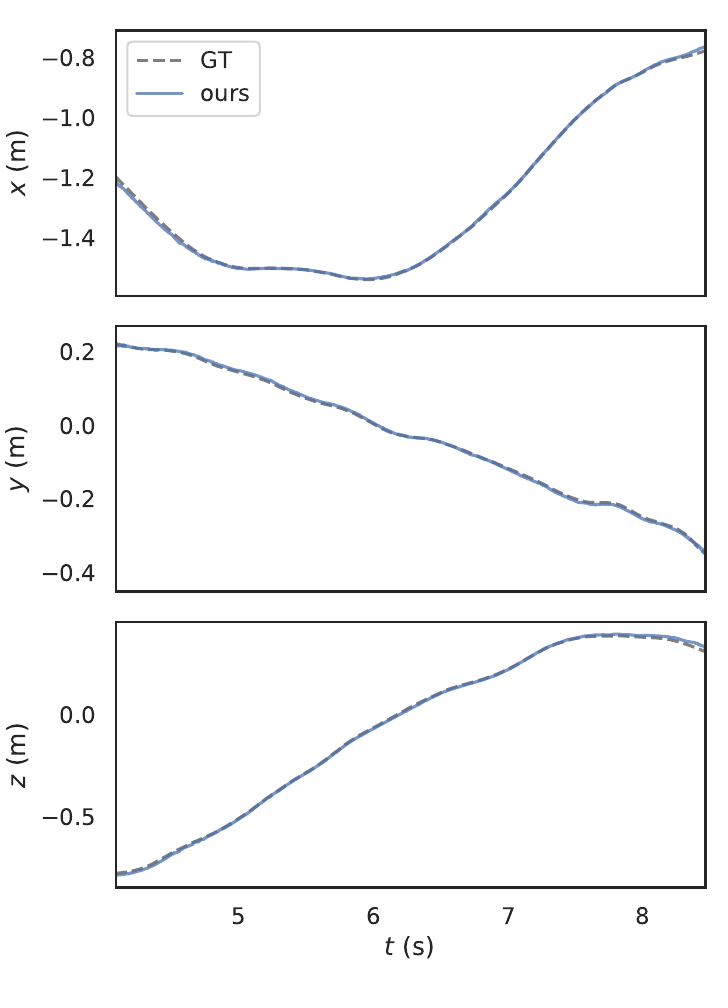}}
    & & 
    % \gframe
    {\includegraphics[trim={0cm 0.5cm 0cm 0.3cm},clip,width=.45\linewidth]{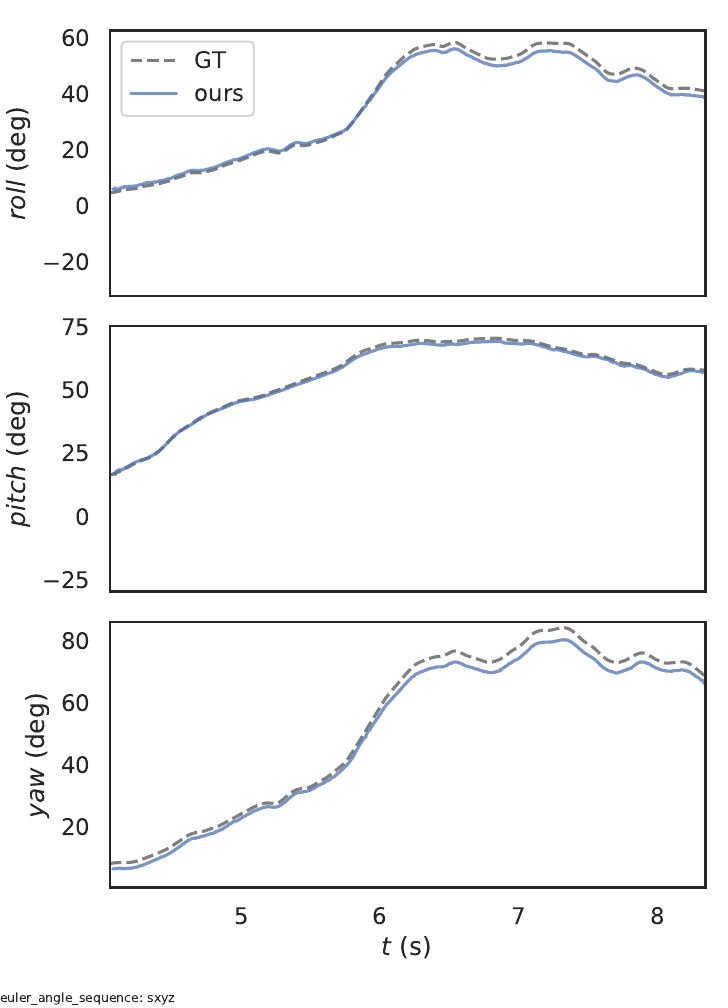}}
\end{tabular}
}
         \caption{Camera Trajectory}
         \label{fig:evimo2:dofs}
     \end{subfigure}
     
    \caption{Results on trinocular camera dataset EVIMO2.
    (a) Sensor rig trajectory, point cloud and depth map from our proposed VO system on \emph{scene\_03\_00\_000000} SFM sequence). 
    The point cloud is color-coded from red (near) to blue (far away).
    (b) Degrees of freedom (DOFs) of estimated poses compared to ground truth poses.}
    \label{fig:evimo2}
\end{figure}

%% file: floats/fig_mvsec.tex
\def\figWidth{0.44\linewidth}
\setlength{\tabcolsep}{1pt}
\begin{figure}[t]
    \centering
    \begin{subfigure}[c]{0.48\linewidth}
         \centering
         \includegraphics[trim={9cm 2cm 5cm 1.5cm},clip,width=.85\linewidth]{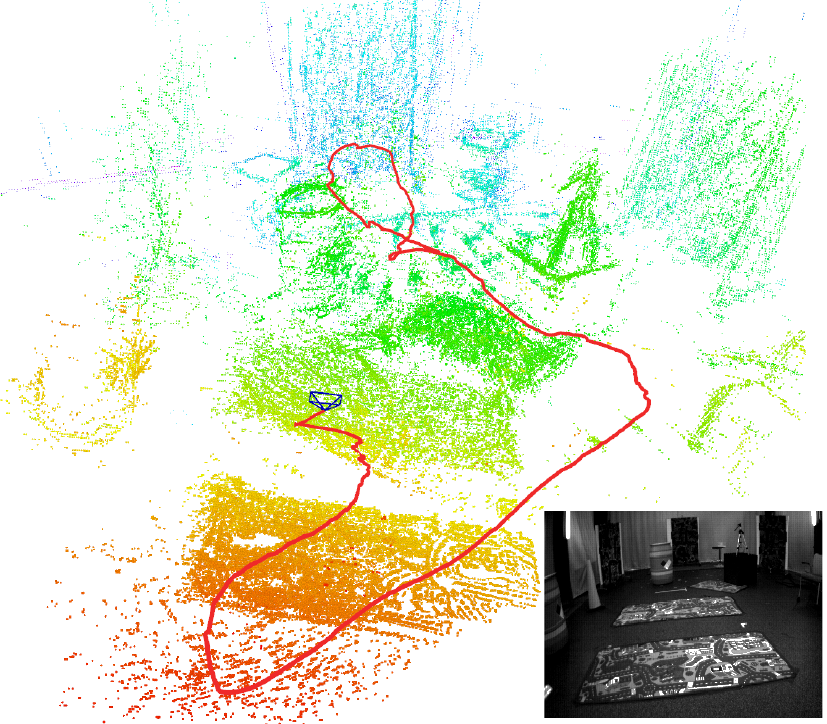}
         \caption{Estimated map and camera trajectory, along with grayscale frame (only for visualization).}
         \label{fig:rviz}
    \end{subfigure}
    \hspace{1ex}
    \begin{subfigure}[c]{0.48\linewidth}
         \centering
{\small
\begin{tabular}{ccc}
    Translation & \, & Rotation \\
    %\gframe
    {\includegraphics[trim={0cm 0.5cm 0cm 0.3cm},clip,width=\figWidth]{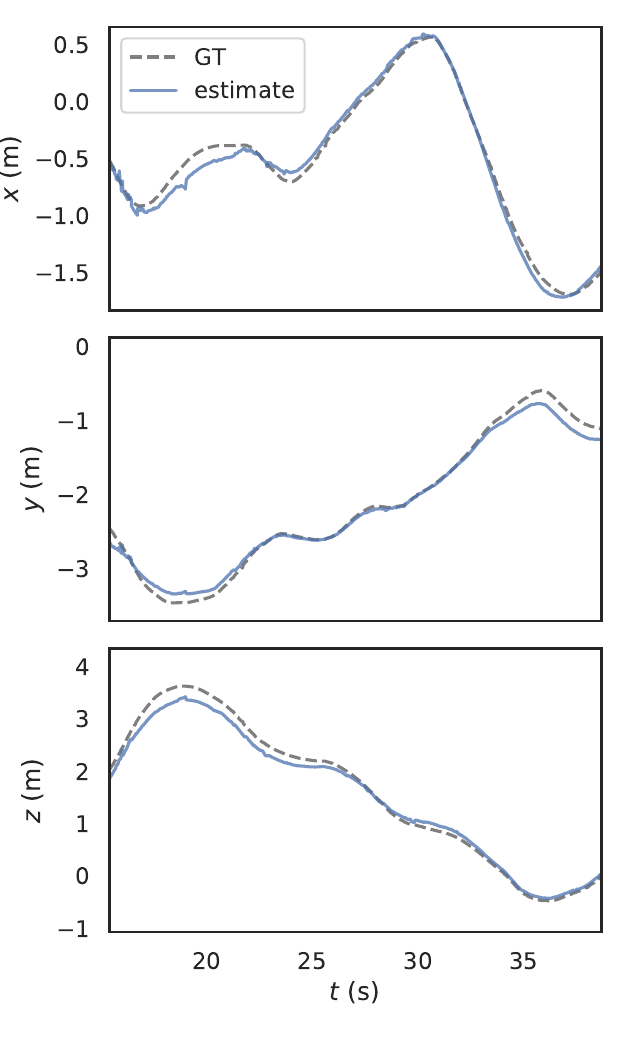}}
    & & 
    %\gframe
    {\includegraphics[trim={0cm 0.5cm 0cm 0.3cm},clip,width=\figWidth]{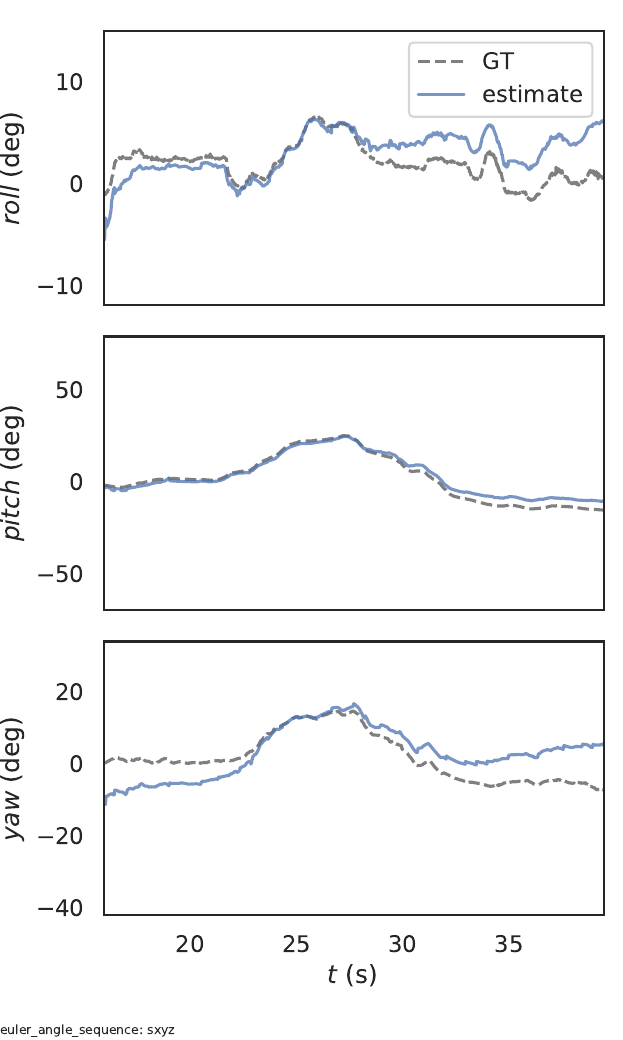}}
\end{tabular}
}
         \caption{Estimated camera poses vs time.}
         \label{fig:dofs}
    \end{subfigure}
    \caption{Results on MVSEC \emph{indoor\_flying1} sequence.}
    \label{fig:mvsec}
\end{figure}

%% file: floats/tab_mapping.tex
\begin{table*}[t]
\caption{Evaluation of depth estimation on DSEC with a maximum range of 50~\si{\meter}. 
All methods use GT poses from LiDAR-inertial odometry.
The methods are evaluated on 35s of stereo data, consisting of 635 million events and containing 350 GT depth maps.
Each depth map is computed using 1~\si{\s} of event data ($\approx10$ million events).
ESVO is executed fusing 10 depth maps generated at 10~\si{\Hz} (LiDAR rate), i.e., 1~\si{\s} of event data. 
}
\label{tab:dsec}
\centering
\begin{adjustbox}{width=\textwidth}
\setlength{\tabcolsep}{4pt}
\begin{tabular}{l*{2}{S[table-format=1.2,table-number-alignment=center]}
*{1}{S[table-format=2.2,table-number-alignment=center]}
*{1}{S[table-format=1.2,table-number-alignment=center]}
*{5}{S[table-format=2.2,table-number-alignment=center]}
*{1}{S[table-format=1.2,table-number-alignment=center]}}
\toprule 
Algorithm & \text{Mean Err} & \text{Median Err} & \text{bad-pix} & \text{SILog Err} & \text{AErrR} & \text{log RMSE} & \text{$\delta < 1.25$} & \text{$\delta < 1.25^2$} & \text{$\delta < 1.25^3$} & \text{\#Points}\\

& \text{[m] $\downarrow$} & \text{[m] $\downarrow$} & \text{[\%] $\downarrow$} & \text{$\times 100 \downarrow$} & \text{[\%] $\downarrow$} & \text{$\times 100 \downarrow$} & \text{[\%] $\uparrow$} & \text{[\%] $\uparrow$} & \text{[\%] $\uparrow$} & \text{[million] $\!\uparrow$}\\
\midrule

EMVS \cite{Rebecq18ijcv} (monocular) & 5.172458115610351 &	0.989746212768555 & 13.963651510274758 &	33.67700376800753 &	59.92076374174398 &	59.97317143274217 &	82.76932186090222 &	87.41417622081045 &	89.14812207502061 &	1.619496\\

ESVO\cite{Zhou20tro} &	3.93158039587316 &	1.624586303710938 & 10.536155257328553 & 8.301632461401524 &	17.655674266315663 &	28.90288447850426 &	84.36975846974358 &	92.80766147539139 &	96.04694242317527 &	\bfseries 9.399507\\
MC-EMVS \cite{Ghosh22aisy} &	3.1318998219085623 &	\bfseries 0.6616444908142096 & 7.681677993030614 &	14.072161706930036 &	24.55673897845241 & 37.848636994042756 & 90.37259064428986 & 93.27839267992007 & 94.84305531449018 &	2.384590\\
Limit rays on MC-EMVS
&	\bfseries 2.8971948175094466 &	0.6750391891479488 & \bfseries 5.973638464663687 &	\bfseries 6.519900668697778 &	\bfseries 11.600603730576153 & \bfseries 25.546060531831655 &	\bfseries 91.54084217490437 &	\bfseries 94.84835567869742 &	\bfseries 96.55983517935124 &	2.311411\\
\bottomrule
\end{tabular}
\end{adjustbox}
\end{table*}

%% file: floats/tab_runtime.tex
\begin{table}[t]
\centering
\caption{Runtime of different steps of our stereo method processing DAVIS346 data.}
\label{tab:runtime}
\begin{adjustbox}{max width=.9\textwidth}
\setlength{\tabcolsep}{6pt}
\begin{tabular}{llc}
\toprule 
 \textbf{Module} & \textbf{Function (Parameters)} & \textbf{Time} [\si{\milli\second}]\\
\midrule
Mapper & DSI creation (2M events per camera, 100 depth planes) & 45\\
& DSI fusion (two DSIs) & 26\\
& arg max & 21\\
& Adaptive Gaussian Thresholding (5$\times$5 px kernel) & 0.2\\
\midrule
Tracker & Event map creation (10k events) & 0.4 \\
& Optimization (1k batch size, 2 pyramid levels, 150 max iterations) & 6--20 \\
\bottomrule
\end{tabular}
\end{adjustbox}
\end{table}